# Towards a Universal Vibration Analysis Dataset: A Framework for Transfer Learning in Predictive Maintenance and Structural Health Monitoring

Mert Sehri[1], Igor Varejão[2], Zehui Hua[1], Vitor Bonella[2], Adriano Santos[3], Francisco de Assis Boldt[3], Patrick Dumond[1] and Flavio Miguel Varejão[2]

[1]*University of Ottawa, Ottawa, ON, K1N 6N5, Canada*

*msehr006@uottawa.ca*
*zhua079@uottawa.ca*
*pdumond@uottawa.ca*

[2]*Universidade Federal do Espírito Santo, Vitoria, ES, 29075-910, Brazil*

*igor.varejao.dev@gmail.com*
*vitor.bonella@edu.ufes.br*
*fvarejao@inf.ufes.br*

[3]*Instituto Federal do Espírito Santo, Vitoria, ES, 29173-087, Brazil*

*adriano_angelo@live.com*
*franciscoa@ifes.edu.br*

**ABSTRACT**

In machine learning (ML), ImageNet has become a reputable resource for transfer learning (TL), allowing the development of efficient ML models with reduced training time and data requirements. However vibration analysis used in fields such as predictive maintenance, structural health monitoring, and fault diagnosis, lacks a comparable large-scale, annotated dataset to facilitate similar advancements. To address this gap, a dataset framework is proposed that begins with a focus on bearing vibration data as an initial step towards creating a universal dataset for vibration-based spectrogram analysis for all machinery. The initial framework includes a collection of bearing vibration signals from various publicly available datasets. To demonstrate the advantages of this framework, experiments were conducted using a deep learning (DL) architecture, showing improvements in model performance when pre-trained on bearing vibration data and fine-tuned on a smaller, domain-specific dataset. These findings highlight the potential to parallel the success of ImageNet in visual computing but for vibration analysis. For future work, this research will include a broader range of vibration signals from multiple types of machinery, emphasizing spectrogram-based representations of the data. Each sample will be labeled according to machinery type, operational status, and the presence or type of faults, ensuring its utility for supervised and unsupervised learning tasks. Additionally, a framework for data preprocessing, feature extraction, and model training specific to vibration data will be developed. This framework will standardize methodologies across the research community, allowing for collaboration and accelerating progress in predictive maintenance, structural health monitoring, and related fields. By mirroring the success of ImageNet in visual computing, this dataset has the potential to improve the development of intelligent systems in industrial applications.

## 1. INTRODUCTION

In recent years, machine learning (ML) has made substantial progress across various fields, driven in large part by the availability of massive, well-annotated datasets (Dean, 2022; Demirbaga, Aujla, Jindal, & Kalyon, 2024). ImageNet, a large-scale dataset widely used in visual computing, is a prime example of how access to large data has allowed for significant advancements in areas like image classification, object detection, and transfer learning (TL) (J. Deng et al., 2009). With the ability to pre-train models on vast and diverse datasets, researchers can fine-tune TL models for more specific tasks with less data and reduced training time, making ML both more efficient and more accessible. However, the success of TL has not yet extended to every domain. In particular, vibration analysis, a crucial field for predictive maintenance, structural health monitoring, and fault diagnosis lacks comparable large-scale datasets (Atmaja, Ihsannur, Suyanto, & Arifianto, 2024). This absence presents a significant barrier to the adoption of ML techniques in TL, where early detection of faults or system degradation can prevent costly failures in industrial applications.

Vibration analysis, unlike visual computing, faces several challenges that complicate the creation of a universal dataset. Machine components often produce complex, non- stationary signals that vary not only by machine type but also by operational conditions, fault types, and environmental factors (Goyal & Pabla, 2016; Tiboni, Remino, Bussola, & Amici, 2022). Another challenge is the significant amount of healthy data when compared to faulty data, since machines aren't meant to fail. Without large, diverse, and labelled datasets, models trained on vibration data tend to lack the generalization ability that is common in visual models trained on datasets like ImageNet (Zhang, Chen, Mao, Zhu, & Xu, 2024). As a result, most of the progress in vibration analysis has been limited to specific domains where small, hand-curated datasets are used (Atmaja et al., 2024). This approach limits the potential for TL, forcing researchers to develop models from scratch for each new application.

To address these issues, a framework is proposed that takes the first step towards developing a comprehensive, large-scale dataset for vibration analysis. The initial focus will be on bearing vibration data, one of the most monitored components in machinery across industries. Bearings are not only critical to the operation of many machines, but their vibration signals provide valuable insights into a machine's health, offering an ideal starting point for this dataset. By curating a broad collection of bearing vibration signals from various public datasets, the aim is to capture a wide range of real-world conditions such as normal operation, and various fault states (inner-race, outer-race, cage, and ball), thereby creating a dataset that is more representative of real-world applications.

In addition to the dataset, the framework proposes leveraging a DenseNet DL architecture (Huang, Liu, Van Der Maaten, & Weinberger, 2017) to demonstrate the effectiveness of TL in bearing diagnosis. By pre-training models on this curated dataset and then fine tuning them on smaller, domain-specific datasets, the hypothesis is that improved model performance can be obtained, even with limited data. This approach would allow for more accurate fault diagnosis, predictive maintenance, and early fault detection, similar to how TL in visual computing has revolutionized image recognition tasks.

Expanding beyond the initial phase, this dataset framework is meant to evolve over time. Future iterations will include not just bearing data but vibration signals from a wide range of machinery, components, types and sensors. Multi-sensor data, such as those captured by accelerometers, and microphones will allow for the development of more sophisticated models that can process different types of signals simultaneously. Such advancements could facilitate sensor fusion techniques, where data from multiple sensors are combined to provide a more comprehensive understanding of machinery health. This will be useful in complex systems, where relying on a single sensor might not provide sufficient information for accurate diagnosis.

Additionally, this framework will incorporate detailed metadata with each sample (input signal length), and fault characteristics of bearing signatures. This level of detail will make the dataset versatile, enabling it to be used in a range of applications from supervised learning tasks where labeled data is abundant, to unsupervised learning and fault detection in cases where labels are scarce. As the dataset grows in scope and complexity, it will position itself as a universal source for vibration analysis, applicable across industries such as manufacturing, aerospace, and energy.

Developing a standardized dataset framework for vibration analysis also holds the potential to unify methodologies across ML research. In the current landscape, the lack of standardization in data collection, preprocessing, and feature extraction leads to inconsistent results and limited collaboration. By introducing a comprehensive, well-annotated dataset accompanied by best practices for data handling and model training, the aim is to encourage a collaborative environment where research can accelerate. Standardization will allow for more reproducibility in experiments and make it easier for researchers to build upon one another's work.

Thus, the proposed dataset framework for vibration analysis represents a crucial step towards addressing the gap in ML resources for TL. By starting with bearing data and expanding to include more types of machinery and sensors, this initiative has the potential to revolutionize how vibration data is analyzed across ML research. By utilizing TL and providing a standardized approach to data handling, this framework has the potential to bring the same level of success to vibration analysis as ImageNet has achieved in visual computing, paving the way for more reliable and intelligent systems in predictive maintenance and machine condition monitoring.

## 2. BACKGROUND

The section focuses on the foundational elements that highlight the requirements to developing a universal vibration dataset framework, including the selection of publicly available bearing datasets, pre-processing methods, and the role of TL in improving fault diagnosis models. By addressing these key components, the creation of a robust and scalable approach to machine condition monitoring can be obtained.

**2.1. Bearing Datasets**

This section introduces the publicly available bearing datasets that are used to create the foundation of the universal dataset for vibration analysis. The selected datasets offer diverse vibration data from multiple bearing types, fault scenarios, and operational conditions, allowing the development of models that can generalize across a wide range of applications. By using publicly available datasets such as the CWRU dataset and combining them with more recent multi-domain datasets like the UORED-VAFCLS, HUST and PADERBORN datasets, this framework will allow for a better approach to TL in fault diagnosis and condition monitoring.

**2.1.1. Case Western Reserve University (CWRU)**

The CWRU dataset is known for its contribution to machine fault diagnosis research. It provides a diverse set of vibration signals such as healthy ball bearings and ball bearings with various fault conditions, including inner race, outer race, and ball faults, collected under different fault severities. This dataset has played a pivotal role in advancing the development of ML models, particularly convolutional neural networks (CNNs), for fault diagnosis tasks.

However, the CWRU dataset was not specifically designed with ML applications in mind, and care must be taken to avoid misleading results. For example, issues related to data with similar load conditions may affect the generalizability of models (Hendriks, Dumond, & Knox, 2022; Rauber, da Silva Loca, Boldt, Rodrigues, & Varejão, 2021). Despite this limitation, it remains a valuable benchmark for the early stages of developing a universal dataset for vibration analysis, especially when data is converted to time-frequency spectrogram images.

The CWRU data consists of sampling frequencies of 12,000 and 48,000 Hz with a sample duration of 10 seconds, and speeds around 1,720 and 1,797 RPM ("Download a Data File | Case School of Engineering | Case Western Reserve University," 2021). The test rig consists of an electrical motor seen on the left-hand side of Figure 1, followed by a shaft with a coupling being tested and a dynamometer to control the different loads being applied. The bearings tested for the CWRU dataset are a SKF 6205-2RS deep groove ball bearing on the drive end, and a SKF 6203-2RS deep groove ball bearing on the fan end.

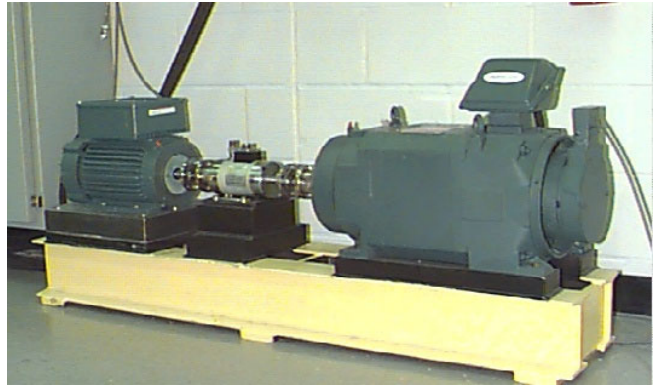

Figure 1. CWRU Bearing Data Test Rig ("Apparatus & Procedures | Case School of Engineering | Case Western Reserve University," 2021)

**2.1.2. University of Ottawa Rolling Element Dataset – Vibration and Acoustic Faults under Constant Load and Speed (UORED-VAFCLS)**

The UORED-VAFCLS dataset complements the CWRU dataset by including vibration and acoustic ball bearing data collected under constant load and speed conditions. This dataset captures various fault types such as ball, cage, inner race, and outer race faults. Additionally, the dataset includes multiple instances of each fault type, sourced from bearings manufactured by different companies, which introduces variability for developing robust TL models.

The inclusion of both vibration and acoustic data under controlled conditions makes the UORED-VAFCLS dataset an excellent resource for multi-sensor fusion, enabling models to learn from diverse input types which will be used in the future of this universal data framework.

The UORED-VAFCLS data consists of a sampling frequency of 42,000 Hz with a sample duration of 10 seconds, and speeds around 1,700 to 1,800 RPM (M. Sehri & Dumond, 2023; Mert Sehri, Dumond, & Bouchard, 2023). The test rig (Figure 2) consists of two bearings tested inside the motor, the bearings tested are the NSK 6203 and FAFNIR 203KD.

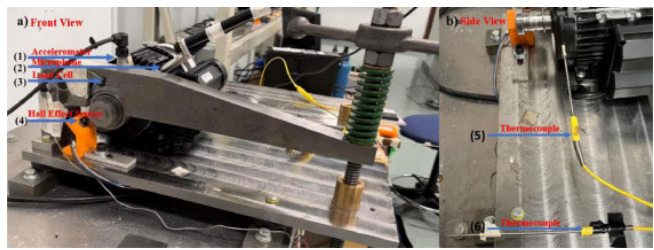

Figure 2. UORED-VAFCLS Bearing Test Rig (Mert Sehri et al., 2023)

### 2.1.3. Hanoi University of Science and Technology (HUST)

The HUST dataset expands on the diversity of bearing sizes and types, offering data from ball bearings such as the 6204, 6205, 6206, 6207, and 6208. Each bearing is tested under multiple fault conditions, including inner race, outer race, and ball faults, which allows for a more comprehensive dataset that supports TL. The variation in bearing size also adds complexity, making it an ideal dataset for developing models that can generalize across different mechanical setups.

Moreover, the HUST dataset can be integrated with the CWRU and UORED-VAFCLS datasets to create a more versatile and extensive dataset. This data fusion approach will enable the development of advanced models that can handle multiple fault types and varying operational conditions.

HUST data consists of a sampling frequency of 25,600 Hz with a sample duration of 20 seconds, and speeds ranging from 600 to 2,100 RPM (Thuan & Hong, 2023). The HUST bearing test rig (Figure 3) consists of mounts for 6204, 6205, 6206, 6207, and 6208 bearings.

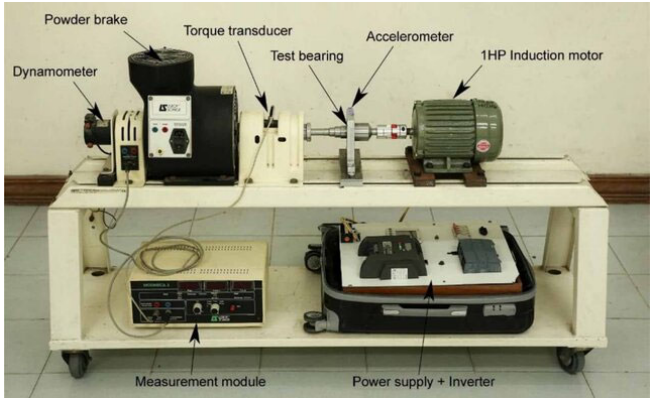

Figure 3. HUST Bearing Data Test Rig

### 2.1.4. PADERBORN Vibration Dataset

The PADERBORN dataset consists of vibration data containing compound fault conditions, captured using three distinct bearing types: deep groove ball bearings, cylindrical roller bearings, and tapered roller bearings. It encompasses data from 21 compound fault scenarios and various rotational speeds, making it a valuable resource for investigating complex machine faults.

This dataset tackles critical challenges faced in industrial environments, such as domain shift and the occurrence of simultaneous faults across multiple components. By incorporating compound faults in both bearings and other rotating components, it extends beyond the limitations of traditional datasets that focus solely on isolated fault conditions.

The PADERBORN dataset includes vibration data recorded using two accelerometers with sampling rates of 8,000 and 16,000 Hz, covering motor speeds of 600, 800, 1,000, 1,200, 1,400, and 1,600 RPM (Y. Deng, 2023; Wang, Wang, Kong, Wang, & Li, 2020). The test rig features 6204 deep groove ball bearings, N204 and NJ204 cylindrical roller bearings, and 30204 tapered roller bearings, making it a robust tool for the study of real-world fault diagnosis challenges. The test rig consists of (1) test motor, (2) measuring shaft, (3) bearing module, (4) flywheel, and (5) load motor seen in Figure 4.

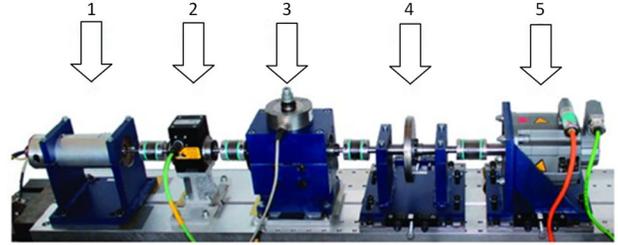

Figure 4. PADERBORN Vibration Data Test Rig (Y. Deng, 2023)

### 2.2. Spectrograms

Spectrograms are designed to capture the varying frequency components of a signal over time. The resulting time-frequency representations (TFRs) are 3D images, where the horizontal axis denotes time, the vertical axis denotes frequency, and the energy of the TFR is represented by different colors. The expression of the resulting TFR based on an original signal $x(t)$ can be written as equation 1.

$$S(\tau, \omega) = \int x(t)g(t-\tau)e^{-i\omega(t-\tau)}\,dt \qquad (1)$$

where $\tau$ and $\omega$ represent time and frequency, $g(t)$ denotes the short window used to truncate the original signal $x(t)$, $i$ is the imaginary unit. Serving as inputs to train the model, spectrogram figures are usually read as RGB figures in three different channels. Also, a simplified version using grayscale images is also widely used since fewer convolutional operations must be performed during training.

Spectrograms are often chosen over statistical preprocessing methods because they provide a rich visual representation of the vibration signal, capturing both time and frequency information in a single domain. This is particularly advantageous for vibration-based intelligent fault diagnosis in industrial applications, where faults often manifest as specific frequency patterns that vary over time. Unlike statistical preprocessing, which summarizes the data into fixed descriptors that may obscure critical fault-related details, spectrograms retain the raw signal's temporal and spectral dynamics, offering a comprehensive view of the data. Furthermore, frequency-based analysis typically outperforms time-based approaches because many fault signatures, such as those caused by imbalance, misalignment, or bearing defects, are more easily identifiable as distinct frequency peaks or patterns. In the time domain, these patterns are often hidden or ambiguous, making it challenging to extract meaningful features. By transforming the data into the frequency or time-frequency domain, spectrograms enable advanced algorithms, especially DL models, to better discern and classify faults, significantly enhancing diagnostic accuracy and reliability in complex industrial settings.

## 2.3. Transfer Learning

TL is a ML technique that involves reusing a pre-trained model on a new but similar problem. There are several common TL techniques, including partial fine tuning and full fine-tuning. In partial fine tuning, the earlier layers of a pre-trained model are used as a fixed feature extractor, meaning that only the final layers of the model are re-trained for the target task. This approach assumes that the learned representations in the lower layers (such as edges and shapes in image processing models) are general enough to apply to the new problem. Full fine-tuning, on the other hand, involves retraining some or all of the model layers on the new task, often with a lower learning rate to preserve previously learned features while adapting to the new data. This approach is beneficial for tasks not closely related to the original training data, allowing the model to better specialize to the specific nuances of the target dataset. Figure 5 provides an overview of the TL process. Using one or more datasets with a large amount of data, a learning technique is applied to create a pre-trained model that captures relevant knowledge about a specific task. Subsequently, a new model can be trained using the pre-trained model along with a smaller dataset for a new task to be performed.

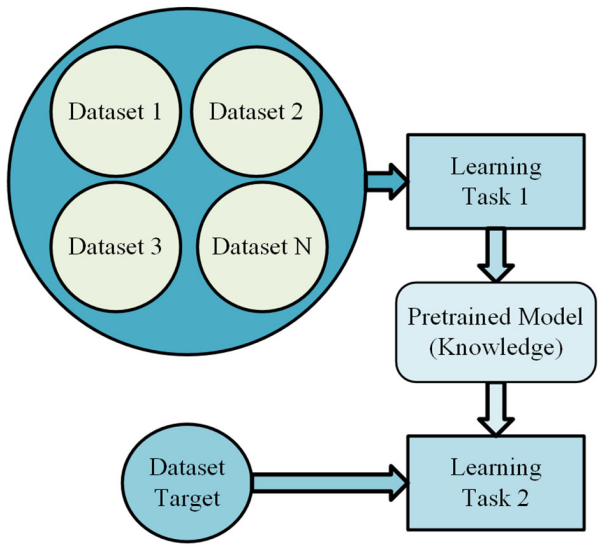

Figure 5. TL Process Flowchart

## 3. VIBNET

VibNet, akin to ImageNet in computer vision, represents a transformative leap for vibration-based intelligent fault diagnosis, addressing critical gaps in research and industry. By providing a comprehensive, structured database of vibration signals from a diverse range of rotating machinery, VibNet facilitates the development, benchmarking, and comparison of advanced diagnostic algorithms. Researchers can use VibNet to access standardized datasets covering various equipment types, fault conditions, and operating environments, enabling more robust and generalizable models. This standardization drives innovation, reducing redundancy in data collection efforts while promoting collaboration and reproducibility across the community.

For industry, VibNet unlocks new possibilities in predictive maintenance and operational efficiency. It empowers engineers to leverage pre-trained DL models and TL techniques tailored to vibration diagnostics, significantly lowering the barrier to implementing AI-driven solutions. By offering a repository of labeled data and associated features, VibNet accelerates the deployment of intelligent systems capable of identifying subtle fault signatures, minimizing downtime, and preventing catastrophic failures. Its role as a unifying platform for data and algorithms not only advances the state of the art but also bridges the gap between academic research and real-world applications, ensuring industries benefit from cutting-edge technologies with reduced development time and cost.

In this section, a preliminary version of VibNet, the dataset framework proposed in this work, is described. It consists of a data storage layer named VibData, a software layer named VibSoft, a VibScript layer containing scripts that describe experiments, and a VibReport layer with reports detailing the results of benchmark methods (i.e., the methods that achieved the best performance in each experiment outlined in VibScript). It is important to highlight that the framework was designed to be extensible, allowing—and indeed intended—that new data, computational software, experimental scripts, and reports be added to the framework at any time. Figure 6 presents a diagram illustrating the layers of VibNet. The diagram illustrates that all layers are encapsulated, ensuring strict separation of responsibilities. Data in the VibData layer can only be accessed through operations provided by the VibSoft layer. Similarly, scripts in the VibScript layer are restricted to utilizing operations defined in the VibSoft layer, and reports in the VibReport layer are exclusively generated using scripts created within the VibScript layer.

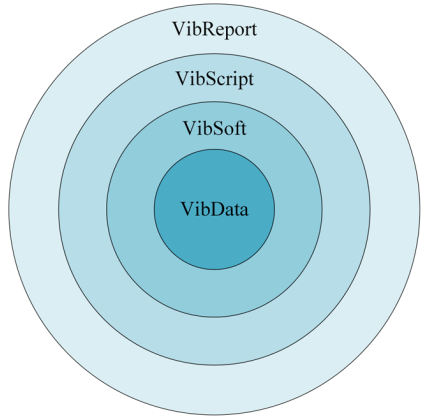

Figure 6. Layer Diagram of VibNet

This section also provides detailed descriptions of the VibData and VibSoft components, and explains the choice of spectrograms as the foundational data used to generate the pre-trained networks in this work. Finally, a detailed description of the TL method employed to create the pre-trained vibration network used in this work's experiments is also provided.

### 3.1. VibData

VibData is a database designed to store and utilize vibration data collected from various types of rotating equipment using different types of sensors. The datasets within VibData are hierarchically organized by equipment type, which may include bearings, gears, motors, compressors, centrifugal pumps, and other rotating machinery. In this preliminary version, only datasets from bearings are used.

Each dataset in VibData contains a collection of data captured from sensors. These data are typically time series representing vibration signal amplitudes measured over a specific time interval at a given sampling frequency. It is also important to note that each time series is collected under specific operating conditions, such as state (healthy, inner race fault, outer race fault, etc.), load, fault severity, sensor positioning, and more. Additionally, each dataset is associated with the type of equipment from which the data were collected, along with their characteristics, such as model, size, and component configuration. Figure 7 provides a macro overview of VibData.

### 3.2. VibSoft

The layer responsible for providing operations on VibData is called VibSoft. It comprises several software libraries designed to enable the following functionalities:

1. **Sampling**: Allows time-domain data to be freely segmented and labeled, creating a dataset of examples.
2. **Resampling**: Enables the example dataset to be split into training, validation, and test sets using various resampling strategies, such as percentage-based splits, cross-validation, and nested cross-validation.
3. **Transformations**: Facilitates the transformation of time-domain data into other domains, such as frequency or time-frequency. This component includes Fourier transforms, wavelets, and routines for generating spectrograms or scalograms.
4. **Feature Extraction**: Provides statistical preprocessing techniques, DL methods, and TL tools to extract features from time-domain or transformed-domain data. These features can be used for training classifiers to diagnose faults.
5. **Feature Selection**: Offers algorithms and techniques to select a more meaningful subset of features for model training. This includes filtering methods or search-based approaches combined with statistical evaluations or wrapper methods.
6. **Fault Detection and Diagnosis**: Includes methods for training classifiers capable of performing fault detection and diagnosis. Techniques for building one-class, binary, multi-class, and multi-label classifiers are available. Notably, this library incorporates TL techniques and tools for generating pre-trained deep vibration networks.

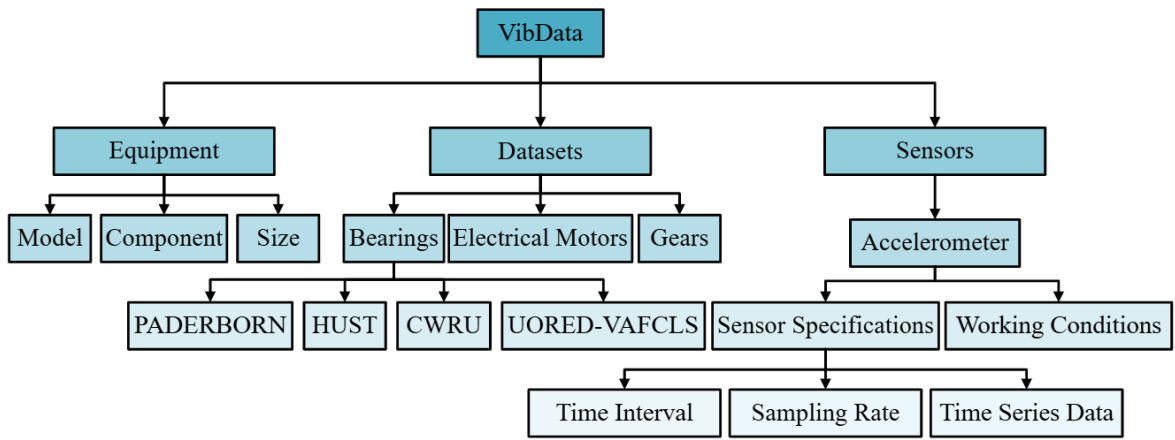

Figure 7. Macro overview of VibData

Each component is designed to ensure flexibility and robustness in handling vibration data for fault detection and diagnosis tasks.

### 3.3. Data Transformation

For the experiments conducted in this preliminary study, the data were transformed into the time-frequency domain by generating spectrograms. This choice was motivated by two factors: (a) frequency-domain analysis is the most commonly used approach for vibration-based fault detection and diagnosis, and (b) spectrograms enable the integration of data obtained with different sampling rates, numbers of sample points, and time intervals. Heterogeneous data collected under varying conditions can be standardized into spectrogram images of a predefined size, ensuring uniformity and compatibility for input into deep neural networks.

The following subsections describe the parameters and values used for generating the spectrograms of the different datasets used in this work.

#### 3.3.1 Parameter settings

To provide clear TFRs from the collected signals, the parameters used for preprocessing the data in preparation for creating spectrograms are listed in Table 1. For each dataset there are 1600 Number of Fast Fourier Transform (NFFT) points, 96% overlap, and a frequency range of 0 to 10 kHz is used.

Table 1. Data Preprocessing Parameters for Spectrograms

| Dataset | Signal length (samples) | Window length | Sampling freq. |
|---|---|---|---|
| CWRU | 12000 | 200 | 48 / 12 kHz |
| UORED-VAFCLS | 10500 | 180 | 42 kHz |
| HUST | 12800 | 200 | 51.2 kHz |
| PADERBORN | 16000 | 180 | 64 kHz |

In Table 1, signal segments with the same time duration are selected for further analysis. Here, 0.25 s is used for all datasets. Also, since the resonance frequency band differs for different experimental setups, a frequency range within [0, 10] kHz is used for the CWRU, UORED-VAFCLS, HUST, and PADERBORN datasets, which ensures that rich fault-related features can be covered in the resulting TFRs. As identified by each experimental test rig, the resonance frequency bands vary with each other, as shown in the Figure 8, where all the Fourier amplitudes versus frequency are drawn. The resonance frequency bands refers to the local peaks that could be observed from the Fourier spectrum. It can be seen that [0, 10 kHz] will cover most of the fault-related impulses for all datasets.

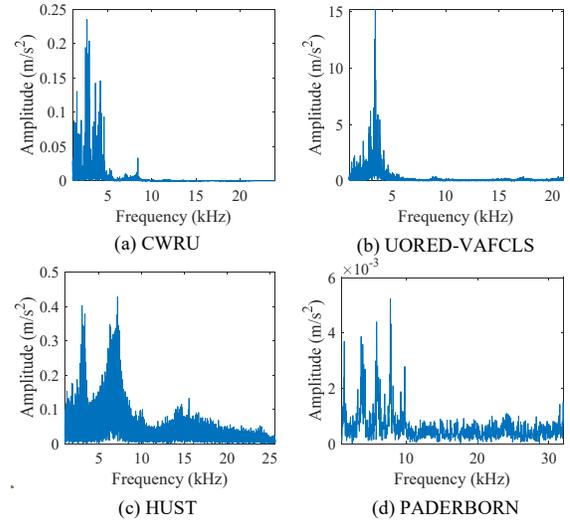

Figure 8. Fourier spectra for different datasets.

By using these parameters, the fault-related impulses can be observed clearly from the spectrograms. It is worth noting that users could also use other parameter combinations as time-frequency analysis has limited time-frequency resolution due to the uncertainty principle. Some examples are given below for illustration, where both the original time domain signal segments and corresponding spectrograms are provided. It can be found that the energy in the obtained spectrograms (Figure 9, 10, 11, and 12) is basically proportional to the amplitude of the collected signals versus time, that is, the larger the amplitude in the original time domain signal waveform, the darker the color in the corresponding spectrograms.

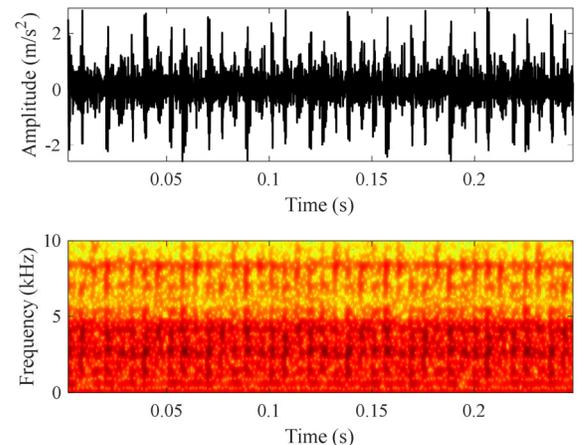

Figure 9. TFR of the CWRU dataset, using the 109.mat file as an example, which has an inner race fault with a size of 0.007 inch under a 0 hp load

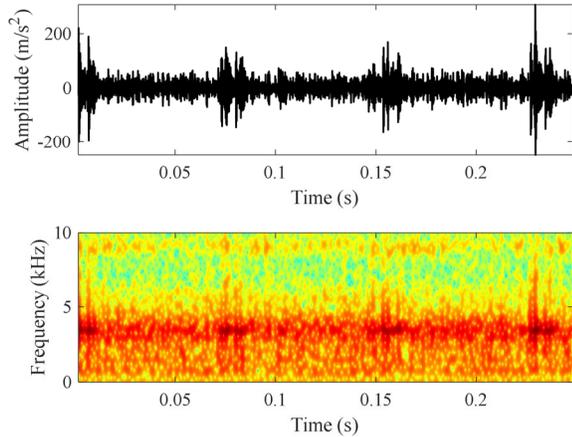

Figure 10. TFR of the UORED-VAFCLS dataset, using the B-11-2 file as an example, which indicates a fully developed ball fault.

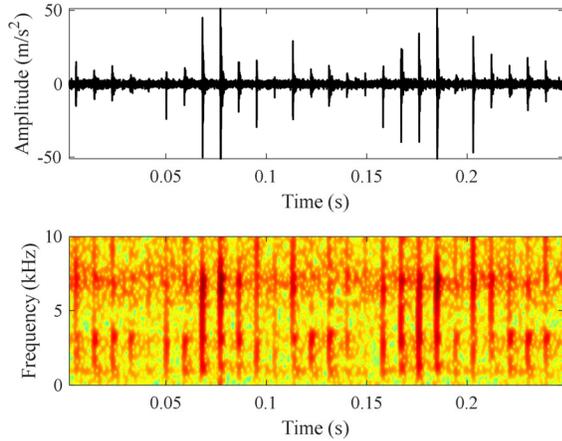

Figure 11. TFR of the HUST dataset, using the B702 file as an example, showing that the 6207 bearing type has a ball fault and a 200W load is applied.

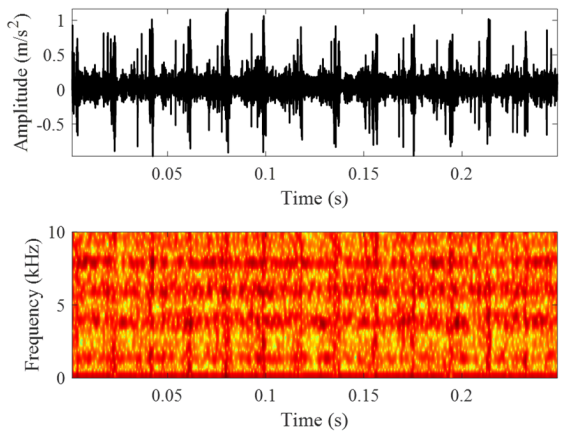

Figure 12. TFR of the Paderborn bearing dataset, using the N09_M07_F10_KI21_9 file, operating parameters are: 900 rpm, 0.7 Nm and 10000 radial force, KI21 indicates damage caused by an accelerated lifetime test, 9 means the 9th trial

### 3.4. Data Evaluation

For this paper, the CWRU dataset is used for testing, where all three accelerometers, located at the drive end (DE), fan end (FE) and base (BA), are included for analysis for both 12kHz and 48kHz, ensuring a great diversity among the evaluated data. K-fold cross validation is used to assess the selected dataset. K-fold cross-validation is a resampling technique used to evaluate ML models by splitting the dataset into k subsets (folds), where the model is trained on a certain number of folds and tested on the remaining fold, repeating this process k times. This method ensures that every data point is used for both training and testing, providing a more robust estimate of model performance compared to a single train-test split. In this work the dataset was divided into folds for training, validation and testing, as seen in Figures 13 and 14.

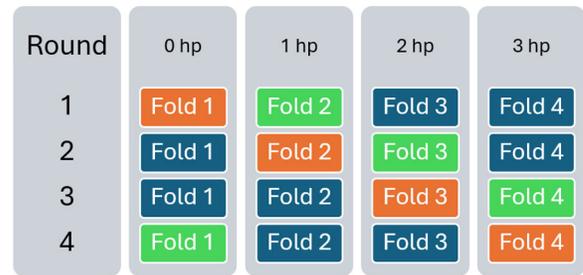

Figure 13. Mitigation bias related to motor load severity

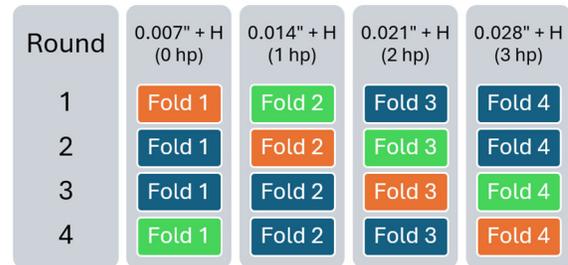

Figure 14. Mitigation bias related to similarity of failure severity

### 3.5. Transfer Learning Method

This work focuses on leveraging TL to improve fault diagnosis using spectrogram-based features and deep neural networks. This section describes the VibNet pre-training pipeline, and the experimental workflow used for evaluating the proposed method.

Figure 15 illustrates the pipeline for creating and using the pre-trained DL network for vibration-based fault diagnosis. Among the four datasets used, the CWRU dataset was selected for evaluating the proposed approach. The CWRU dataset was chosen for resampling analysis as it was used to test the different approaches, ensuring a consistent evaluation framework. Figure 15 shows the process of creating the VibNet pre-trained deep network. First the HUST, UORED-VAFCLS, and PADERBORN datasets are sampled to generate the examples used for training VibNet. Then these samples are used for generating corresponding spectrogram images of the examples. In this study, the CWRU dataset is subjected to the same process used for training VibNet. Instead, CWRU spectrogram examples are resampled for composing the CWRU training, validation and testing datasets used for evaluating the models used in this work.

The other datasets were primarily used for training VibNet, which is why they were not included in this specific resampling analysis. It is worth noting that this pipeline was evaluated using two different data divisions, as detailed in Section 3.4. The first division uses load condition information from the CWRU dataset, while the second focuses on fault size. This approach ensures multiple evaluation modes, strengthening the robustness of our method.

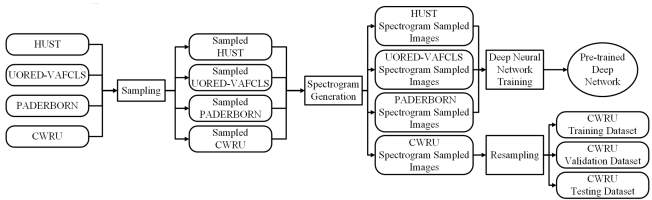

Figure 15. TL Experimental Workflow

Here's a summary of the pipeline's key steps:

1. **Datasets and Sampling**: Four datasets are used: HUST, UORED-VAFCLS, PADERBORN, and CWRU. CWRU serves as a benchmark to evaluate the various methods implemented, while HUST, UORED-VAFCLS, and PADERBORN are employed to pre-train VibNet, in contrast to ImageNet. For each training, the records from each dataset are segmented into temporal windows to be processed into the time-frequency spectrum.
2. **Spectrogram Generation**: The sampled data from all datasets are converted into spectrogram images, a representation of the vibration signals in the time-frequency domain are provided in section 3.3.1. Given the difference in scales between the datasets and the need to standardize the size of the images to be fed into the neural network, it was necessary to resize each image to a fixed size. This fixed value was 256x512 due to the lower compression and consequently less loss of information in the resulting spectra.
3. **Pre-training the Deep Network**: The spectrogram images from the HUST, UORED-VAFCLS, and PADERBORN datasets are used to train the deep neural network. The chosen architecture is DenseNet121, which has demonstrated excellent performance in the field of computer vision, such as on the CIFAR-10 classification benchmark and ImageNet. It helps alleviate the vanishing gradient problem, strengthens feature propagation, encourages feature reuse, and offers several other benefits. With this set, it was possible to use both the architecture from scratch and the pre-trained model from ImageNet

Figure 16 shows the workflow used for training and evaluating DL models for fault diagnosis using vibration data from the CWRU dataset. In this work, the code for data management vibdata and the experiments vibnet-experiments is publicly available on GitHub in two separate repositories. This separation ensures better modularity and organization of the experiments. By providing the code, we enable future researchers to build upon and extend this work, fostering innovation and allowing for the exploration of new ideas and improvements. With this set, the workflow for the experiments includes the following key steps:

1. **Pre-trained Model**: A pre-trained deep network is used from ImageNet. The **DenseNet121 pre-trained model** is used. Following that the pretraining of VibNet is performed using the DesneNet121 topology, but without importing parameter weights. The training is done with the source datasets (Hust, UORED-VAFCLS and PADEBORN).
2. **CWRU Dataset Partitioning**: The CWRU spectrogram images undergo resampling to create training, validation, and testing datasets as detailed in Section 3.4.
3. **Fine-Tuning**: Both pre-trained networks are fine-tuned using the CWRU training and validation dataset. At this stage, we employed two approaches. In the first, we update all the weights using the CWRU dataset, known as full fine tuning. In the second, we only trained the classification header, while keeping the feature extractor weights frozen, called partial fine tuning.
4. **Evaluation**: The models are (VibNet fine-tuned, ImageNet fine-tuned, and from-scratch trained) tested on the CWRU testing dataset. To address the class imbalance in the target dataset, two metrics more suited to this situation were used: balanced accuracy and F1-macro, both of which provide a better evaluation of performance across imbalanced classes.

5. **Results**: The testing phase generates results that are used to compare the performance of these approaches in terms of their effectiveness for fault diagnosis.

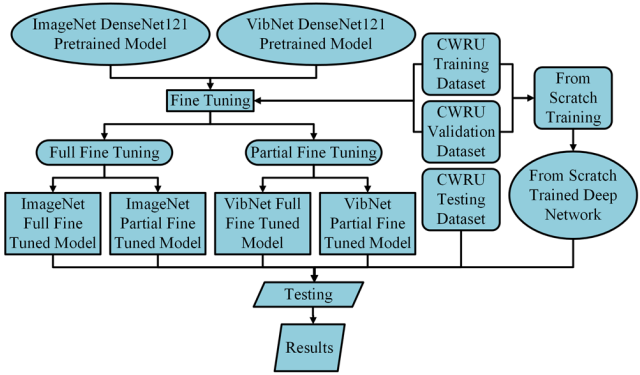

Figure 16. TL Experimental Workflow

For training the DenseNet121 model, an EarlyStopping callback was used to prevent overfitting. Additionally, a ReduceLROnPlateau learning rate scheduler was implemented to adjust the learning rate when the validation loss plateaued. A batch size of 32 was used across all experiments, while the number of epochs and initial learning rate were varied to optimize model convergence, based on validation performance. These hyperparameters are summarized in Table 2.

Table 2. Experiment Hyperparemeters

| Type | Experiment | Epochs | Initial Learning Rate |
|---|---|---|---|
| CWRU By Load | Partial-Fine-Tuning-ImageNet | 50 | 0.01 |
| | Partial-FineTuning-VibNet | 50 | 0.01 |
| | Full-FineTuning-ImageNet | 15 | 0.001 |
| | Full-FineTuning-VibNet | 25 | 0.01 |
| | From-Scratch | 15 | 0.001 |
| CWRU by Fault Severity | Partial-Fine-Tuning-ImageNet | 50 | 0.001 |
| | Partial-FineTuning-VibNet | 50 | 0.01 |
| | Full-FineTuning-ImageNet | 25 | 0.001 |
| | Full-FineTuning-VibNet | 25 | 0.01 |
| | From-Scratch | 25 | 0.001 |

This setup evaluates the benefits of TL using a VibNet pre-trained model comparing the proposed model with two common intelligent fault diagnosis DL-based method: training a deep network from scratch and by using TL with an ImageNet pre-trained network.

## 4. RESULTS & DISCUSSION

This section demonstrates the potential of VibNet by displaying vibration data results when training multiple datasets.

### 4.1. Data Transformation into Spectrograms

As depicted in Section 3.4, the CWRU dataset is used for testing, while the other 3 datasets are used for training. The vibration signals from the selected 4 datasets are transformed into spectrograms using the short-time Fourier transform (STFT) using a Hanning window.

### 4.2. Data Organization and Sample Distribution

To facilitate the analysis and ensure structured experimentation, the samples from the CWRU dataset were organized into directories based on two experimental configurations: load conditions and fault severity (Table 3). The remaining breakdown of the number of samples used for testing is shown in Table 3, where H, I, O and B represent healthy, inner race, outer race and ball faults, respectively.

Table 3. Number of Samples by Bearing State

| Dataset | Folds | H | I | O | B |
|---|---|---|---|---|---|
| CWRU by Loads | Fold 1 (0 hp) | 40 | 270 | 280 | 290 |
| | Fold 2 (1 hp) | 80 | 412 | 420 | 430 |
| | Fold 3 (2 hp) | 80 | 430 | 420 | 430 |
| | Fold 4 (3 hp) | 80 | 430 | 420 | 430 |
| | Total | 280 | 1542 | 1540 | 1580 |
| CWRU by Fault Severity | Fold 1 (0.007") | 40 | 520 | 520 | 520 |
| | Fold 2 (0.014") | 80 | 462 | 520 | 520 |
| | Fold 3 (0.021") | 80 | 500 | 500 | 520 |
| | Fold 4 (0.028") | 80 | 40 | 0 | 40 |
| | Total | 280 | 1522 | 1540 | 1600 |
| HUST | N/A | 600 | 600 | 600 | 405 |
| UORED-VAFCLS | N/A | 800 | 200 | 200 | 200 |
| PADERBORN | N/A | 1918 | 1267 | 1907 | 0 |

### 4.3. Results and Discussion

The results for K-fold cross-validation (K = 4) are presented in Table 4, highlighting the mean and standard deviation of balanced accuracy and macro F1 scores for various approaches, including full fine tuning, partial fine tuning, and training from scratch. The best results in the table are indicated in bold.

For the CWRU Load Division, the full fine tuning method from ImageNet weights achieved a mean balanced accuracy of 97.58% and an macro F1 score of 97.46%, with standard deviations of 4.30% and 4.53%, respectively. In comparison, Full fine tuning from VibNet weights resulted in a slightly lower mean balanced accuracy of 97.39% and a macro F1 score of 97.32%, but with reduced standard deviations of 3.79% and 3.94%. Partial fine tuning on VibNet weights showed a mean balanced accuracy of 95.66% and a macro F1 score of 95.66%, while partial fine tuning on ImageNet weights resulted in significantly lower performance, with a mean accuracy of 90.04% and an F1 score of 89.94%.

In the CWRU Severity Division, the full fine tuning method on ImageNet weights achieved a mean accuracy of 64.38% and an F1 score of 65.12%, with higher standard deviations of 21.39% and 21.69%. Full fine tuning on VibNet weights improved the mean accuracy to 66.57% and the F1 score to 65.91%, albeit with slightly higher standard deviations of 22.47% and 22.94%. The training from scratch model resulted in a mean accuracy of 64.35% and an F1 score of 58.00%, with a notable reduction in the standard deviation of the F1 score to 10.43%. Partial fine tuning on VibNet weights showed the best performance in this division, with a mean accuracy of 66.88% and an F1 score of 66.21%, while Partial fine tuning on ImageNet weights had the lowest performance, with a mean accuracy of 58.26% and an F1 score of 52.75%. The results of these preliminary experiments were quite satisfactory. VibNet full tuning achieved the second-best results and equivalent to the best ImageNet full fine tuning on the load division and VibNet partial and full fine tuning respectively achieved the best and second-best results in the severity division, which is the more difficult diagnosis problem in the CWRU dataset.

Table 4. ImageNet and VibNet Dataset Results

| Metric | Method | Mean (%) | | Std (%) | |
|---|---|---|---|---|---|
| | | Accuracy Score | F1 Score | Accuracy Score | F1 Score |
| CWRU Load Division | Full Finetune ImageNet | **97.58** | **97.46** | 4.30 | 4.53 |
| | Full Finetune VibNet | 97.39 | 97.32 | 3.79 | 3.94 |
| | From Scratch | 96.84 | 96.81 | 3.82 | 3.89 |
| | Partial Finetune VibNet | 95.66 | 95.66 | **3.62** | **3.63** |
| | Partial Finetune ImageNet | 90.04 | 89.94 | 14.86 | 15.53 |
| CWRU Severity Division | Full Finetune ImageNet | 64.38 | 65.12 | 21.39 | 21.69 |
| | Full Finetune VibNet | 66.57 | 65.91 | 22.47 | 22.94 |
| | From Scratch | 64.35 | 58.00 | 20.18 | 10.43 |
| | Partial Finetune VibNet | **66.88** | **66.21** | 21.38 | 21.69 |
| | Partial Finetune ImageNet | 58.26 | 52.75 | **4.31** | **7.00** |

## 5. CONCLUSION

Developing a standardized dataset framework for vibration analysis is an essential step toward bridging the current gap in ML applications, particularly for TL. Starting with the curation of bearing vibration data, the VibNet framework aims to provide a diverse dataset that can aid in identifying the health conditions of bearings. By pre-training models on this robust dataset and fine tuning on smaller bearing datasets, the goal is to enhance model generalization and performance, even when there is a limited amount of data. This will allow for effective fault diagnosis, predictive maintenance, and in addressing challenges that limit the advancement of ML in vibration analysis for specific domains. Experiments performed in this work present promising results. Even when training with a small number of vibration datasets, the VibNet pre-trained model achieved better results than state of the art fault diagnosis methods, particularly with the consolidated ImageNet transfer learning approach. The addition of more bearing datasets is expected to help improve results further.

Aside from bearing data, the future of the VibNet framework is designed to evolve and include data from various machinery, potentially enabling multi-sensor analysis and advanced sensor fusion techniques. The inclusion of detailed metadata and diverse input conditions will make the datasets applicable across different ML tasks, from supervised learning to unsupervised fault detection. By presenting a standardization in data handling (preprocessing), the proposed framework can unify ML methodologies, promoting reproducibility and collaborative growth. The goal is to set the stage so that ML-based vibration analysis can reach a level of progress similar to what ImageNet has achieved for visual computing, unlocking more reliable and intelligent systems for machine condition monitoring and predictive maintenance across industries. Future research will focus on the broader validation of different machinery components across real world scenarios.

## NOMENCLATURE

DL    deep learning
ML    machine learning
TL    transfer learning
CNN    convolutional neural network

## REFERENCES

Apparatus & Procedures | Case School of Engineering | Case Western Reserve University. (2021, August 10). Retrieved February 16, 2023, from Case School of Engineering website: https://engineering.case.edu/bearingdatacenter/apparatus-and-procedures

Atmaja, B. T., Ihsannur, H., Suyanto, & Arifianto, D. (2024). Lab-Scale Vibration Analysis Dataset and Baseline Methods for Machinery Fault Diagnosis with Machine Learning. *Journal of Vibration Engineering & Technologies*, *12*(2), 1991–2001. https://doi.org/10.1007/s42417-023-00959-9

Dean, J. (2022). A Golden Decade of Deep Learning: Computing Systems & Applications. *Daedalus*, *151*(2), 58–74. https://doi.org/10.1162/daed_a_01900

Demirbaga, Ü., Aujla, G. S., Jindal, A., & Kalyon, O. (2024). Machine Learning for Big Data Analytics. In Ü. Demirbaga, G. S. Aujla, A. Jindal, & O. Kalyon (Eds.), *Big Data Analytics: Theory, Techniques, Platforms, and Applications* (pp. 193–231). Cham: Springer Nature Switzerland. https://doi.org/10.1007/978-3-031-55639-5_9

Deng, J., Dong, W., Socher, R., Li, L.-J., Li, K., & Fei-Fei, L. (2009). ImageNet: A large-scale hierarchical image database. *2009 IEEE Conference on Computer Vision and Pattern Recognition*, 248–255. https://doi.org/10.1109/CVPR.2009.5206848

INTERNATIONAL JOURNAL OF PROGNOSTICS AND HEALTH MANAGEMENTDeng, Y. (2023). *Paderborn bearing dataset and PHM2009 gearbox dataset*. 1. https://doi.org/10.17632/65d3pzth7v.1

Download a Data File | Case School of Engineering | Case Western Reserve University. (2021, August 10). Retrieved January 15, 2024, from Case School of Engineering website: https://engineering.case.edu/bearingdatacenter/download-data-file

Goyal, D., & Pabla, B. S. (2016). The Vibration Monitoring Methods and Signal Processing Techniques for Structural Health Monitoring: A Review. *Archives of Computational Methods in Engineering*, *23*(4), 585–594. https://doi.org/10.1007/s11831-015-9145-0

Hendriks, J., Dumond, P., & Knox, D. A. (2022). Towards better benchmarking using the CWRU bearing fault dataset. *Mechanical Systems and Signal Processing*, *169*, 108732. https://doi.org/10.1016/j.ymssp.2021.108732

Huang, G., Liu, Z., Van Der Maaten, L., & Weinberger, K. Q. (2017). Densely Connected Convolutional Networks. *2017 IEEE Conference on Computer Vision and Pattern Recognition (CVPR)*, 2261–2269. https://doi.org/10.1109/CVPR.2017.243

Rauber, T. W., da Silva Loca, A. L., Boldt, F. de A., Rodrigues, A. L., & Varejão, F. M. (2021). An experimental methodology to evaluate machine learning methods for fault diagnosis based on vibration signals. *Expert Systems with Applications*, *167*, 114022. https://doi.org/10.1016/j.eswa.2020.114022

Sehri, M., & Dumond, P. (2023). University of Ottawa rolling-element dataset–vibration and acoustic faults under constant load and speed conditions (UORED-VAFCLS). *Vol*, *5*, 10–17632.

Sehri, Mert, Dumond, P., & Bouchard, M. (2023). University of Ottawa constant load and speed rolling-element bearing vibration and acoustic fault signature datasets. *Data in Brief*, *49*, 109327. https://doi.org/10.1016/j.dib.2023.109327

Thuan, N. D., & Hong, H. S. (2023). HUST bearing: A practical dataset for ball bearing fault diagnosis. *BMC Research Notes*, *16*(1), 138. https://doi.org/10.1186/s13104-023-06400-4

Tiboni, M., Remino, C., Bussola, R., & Amici, C. (2022). A Review on Vibration-Based Condition Monitoring of Rotating Machinery. *Applied Sciences*, *12*(3), 972. https://doi.org/10.3390/app12030972

Wang, S., Wang, D., Kong, D., Wang, J., & Li, W. (2020). Few-Shot Rolling Bearing Fault Diagnosis with Metric-Based Meta Learning. *Sensors*, *20*, 6437. https://doi.org/10.3390/s20226437

Zhang, T., Chen, H., Mao, X., Zhu, X., & Xu, L. (2024). A Domain Generation Diagnosis Framework for Unseen Conditions Based on Adaptive Feature Fusion and Augmentation. *Mathematics*, *12*(18), 2865. https://doi.org/10.3390/math12182865
12